\documentclass{article}

\usepackage[utf8]{inputenc}
\usepackage{microtype}
\usepackage{graphicx}
\usepackage{subfigure}
\usepackage{booktabs} % for professional tables
\usepackage{lipsum}

\usepackage{amsmath}
\usepackage{amssymb}

\usepackage{pifont}
\usepackage{xfrac}

\newcommand{\cmark}{\ding{51}}%
\newcommand{\xmark}{\ding{55}}%

\newcommand{\eos}{\langle \mathrm{EOS} \rangle}

\DeclareMathOperator*{\argmax}{argmax}

\usepackage{xcolor}
\definecolor{lightgray}{gray}{0.7}

\usepackage{ulem} % Underlined text
\usepackage{tikz}
\usetikzlibrary{positioning}

\usepackage{pgfplots}
\pgfplotsset{compat=1.14}

\usepackage{hyperref}

\usepackage[preprint]{icml2019}

\icmltitlerunning{Insertion Transformer: Flexible Sequence Generation via Insertion Operations}

\begin{document}

\twocolumn[
\icmltitle{Insertion Transformer: \\ Flexible Sequence Generation via Insertion Operations}
% Flexible Sequence Generation via Insertion Operations
% Ordered and Unordered Sequence Generation
% Order vs Chaos



\icmlsetsymbol{equal}{*}

\begin{icmlauthorlist}
\icmlauthor{Mitchell Stern}{google,berkeley}
\icmlauthor{William Chan}{google}
\icmlauthor{Jamie Kiros}{google}
\icmlauthor{Jakob Uszkoreit}{google}
\end{icmlauthorlist}

\icmlaffiliation{google}{Google Brain, Mountain View, Toronto, Berlin}
\icmlaffiliation{berkeley}{University of California, Berkeley}

\icmlcorrespondingauthor{Mitchell Stern}{\mbox{mitchell@berkeley.edu}}
\icmlcorrespondingauthor{William Chan}{\mbox{williamchan@google.com}}

\icmlkeywords{Machine Learning, Machine Translation}

\vskip 0.3in
]



\printAffiliationsAndNotice{}  % leave blank if no need to mention equal contribution

\begin{abstract}
We present the Insertion Transformer, an iterative, partially autoregressive model for sequence generation based on insertion operations. Unlike typical autoregressive models which rely on a fixed, often left-to-right ordering of the output, our approach accommodates arbitrary orderings by allowing for tokens to be inserted anywhere in the sequence during decoding. This flexibility confers a number of advantages: for instance, not only can our model be trained to follow specific orderings such as left-to-right generation or a binary tree traversal, but it can also be trained to maximize entropy over all valid insertions for robustness. In addition, our model seamlessly accommodates both fully autoregressive generation (one insertion at a time) and partially autoregressive generation (simultaneous insertions at multiple locations). We validate our approach by analyzing its performance on the WMT 2014 English-German machine translation task under various settings for training and decoding. We find that the Insertion Transformer outperforms many prior non-autoregressive approaches to translation at comparable or better levels of parallelism, and successfully recovers the performance of the original Transformer while requiring only logarithmically many iterations during decoding.
\end{abstract}

\section{Introduction}

\begin{figure*}[t]
\small
\hfill
\begin{minipage}[t]{0.44\textwidth}
\centering
\vspace{0pt}
{\normalsize Serial generation:} \\
\vspace{0.5em}
\begin{tabular}{cll}
\toprule
$t$ & Canvas & Insertion \\
\midrule
0 & [] & $(\text{ate}, 0)$ \\
1 & [\textcolor{blue}{\uline{ate}}] & $(\text{together}, 1)$ \\
2 & [ate, \textcolor{blue}{\uline{together}}] & $(\text{friends}, 0)$ \\
3 & [\textcolor{blue}{\uline{friends}}, ate, together] & $(\text{three}, 0)$ \\
4 & [\textcolor{blue}{\uline{three}}, friends, ate, together] & $(\text{lunch}, 3)$ \\
5 & [three, friends, ate, \textcolor{blue}{\uline{lunch}}, together] & $(\eos, 5)$ \\
\end{tabular}
\end{minipage}
\hfill
\begin{minipage}[t]{0.53\textwidth}
\centering
\vspace{0pt}
{\normalsize Parallel generation:} \\
\vspace{0.5em}
\begin{tabular}{cll}
\toprule
$t$ & Canvas & Insertions \\
\midrule
0 & [] & $(\text{ate}, 0)$ \\
1 & [\textcolor{blue}{\uline{ate}}] & $(\text{friends}, 0)$, $(\text{together}, 1)$ \\
2 & [\textcolor{blue}{\uline{friends}}, ate, \textcolor{blue}{\uline{together}}] & $(\text{three}, 0)$, $(\text{lunch}, 2)$ \\
3 & [\textcolor{blue}{\uline{three}}, friends, ate, \textcolor{blue}{\uline{lunch}}, together] & $(\eos, 5)$ \\
\end{tabular}
\end{minipage}
\hfill
\caption{Examples demonstrating how the clause ``three friends ate lunch together'' can be generated using our insertion framework. On the left, a serial generation process is used in which one insertion is performed at a time. On the right, a parallel generation process is used with multiple insertions being allowed per time step. Our model can either be trained to follow specific orderings or to maximize entropy over all valid actions. Some options permit highly efficient parallel decoding, as shown in our experiments.}
\label{fig:example-basic}
\end{figure*}

Neural sequence models \cite{sutskever-nips-2014,cho-emnlp-2014} have been successfully applied to many applications, including machine translation \cite{bahdanau-iclr-2015,luong-emnlp-2015}, speech recognition \cite{bahdanau-icassp-2016,chan-icassp-2016}, speech synthesis \cite{oord-arxiv-2016,wang-interspeech-2017}, image captioning \cite{vinyals-cvpr-2015,xu-icml-2015} and image generation \cite{oord-icml-2016,oord-nips-2016}. These models have a common theme: they rely on the chain-rule factorization and have an autoregressive left-to-right structure. This formulation bestows many advantages in both training and inference. Log-likelihood computation is tractable, allowing for efficient maximum likelihood learning. Efficient approximate inference is also made possible through beam search decoding. However, the autoregressive framework does not easily accommodate for parallel token generation or more elaborate generation orderings (e.g., tree orders).

More recently, there has been work on non-autoregressive sequence models such as the Non-Autoregressive Transformer (NAT) \cite{gu-iclr-2018} and the Iterative Refinement model \cite{lee-emnlp-2018}.
% In both of these works, the first step is to implicitly (via fertilities \cite{gu-iclr-2018}) or explicitly \cite{lee-emnlp-2018} model the target sequence length given the source sequence. Given the target sequence length, a modified copy of the source sequence is created via fertilities \cite{gu-iclr-2018}, or truncation and padding \cite{lee-emnlp-2018}. The decoder conditions on this modified source sequence, and produces the target sequence in parallel. \citet{lee-emnlp-2018} adds a second decoder iterative refinement stage, wherein a new target sequence hypothesis is produced conditioned on the previous target sequence hypothesis.
In both of these models, the decoder is seeded with an initial input derived from the source sequence, then produces the entire target sequence in parallel. \citet{lee-emnlp-2018} adds an iterative refinement stage to the decoder in which a new hypothesis is produced conditioning on the input and the previous output.

While allowing for highly parallel generation, there are a few drawbacks to such approaches.
% The target sequence length needs to be modelled first, this fixes the tar-get sequence length and does not allow the target sequence length to dynamically grow afterward
The first is that the target sequence length needs to be chosen up front, preventing the output from growing dynamically as generation proceeds. This can be problematic if the chosen length is too short to accommodate the desired target, or can be wasteful if it is too long. In the case of \citet{gu-iclr-2018}, there is also a strong conditional independence assumption between output tokens, limiting the model's expressive power. \citet{lee-emnlp-2018} relaxes this assumption but in turn requires two separate decoders for the initial hypothesis generation and the iterative refinement stage.
% In the case of \citet{gu-iclr-2018}, this requires relying on a phrased-based translation model \cite{dyer-naacl-2013} to model the fertilities \cite{brown-cl-1993}.

In this work, we present a flexible sequence generation framework based on insertion operations.
The Insertion Transformer is an iterative, partially autoregressive model which can be trained in a fully end-to-end fashion.
Generation is accomplished by repeatedly making insertions into an initially-empty output sequence until a termination condition is met.
%Flexiblity exists in both training and inference. During training, the Insertion Transformer can be optimized for a left-to-right order, balanced binary tree order, or even learning all possible orders. While for inference, Insertion Transformer can be decoded serially one insertion at a time, or in parallel with simultaneous insertions at multiple locations.
Our approach bypasses the problem of needing to predict the target sequence length ahead of time by allowing the output to grow dynamically, and also permits deviation from classic left-to-right generation, allowing for more exotic orderings like balanced binary trees.
%First, we bypass the problem of needing to predict the the target sequence length beforehand, by adopting a formulation that allows our target sequence to dynamically grow in size. Starting from an empty sequence, the sequence is grown dynamically through a series of insertion operations. At each iteration, the model conditions on the previous sequence hypothesis and produces the next set of insertion operations to modify the sequence. The insertion operations need not follow a left-to-right generation order, and can follow complex orderings such as a balanced binary tree order, or even be trained to learn all possible orderings.

During inference, the Insertion Transformer can be used in an autoregressive manner for serial decoding, with one insertion operation being applied at a time, or in a partially autoregressive manner for parallel decoding, with insertions at multiple locations being applied simultaneously. This allows for the target sequence to grow exponentially in length. In the case of a balanced binary tree order, our model can use as few as $\lfloor\log_2 n\rfloor + 1$ operations to produce a sequence of length $n$, which we find achievable in practice using an appropriately chosen loss function during training.
%generate a sequence of length $n$ in $O(\log n)$ iterations.

% Our Insertion Transformer model an end-to-end model trained in an end-to-end fashion. We do not require a separate phrase-based translation model, nor do we require multiple decoders.

\section{Sequence Generation via Insertion Operations}

In this section, we describe the abstract framework used by the Insertion Transformer for sequence generation. The next section then describes the concrete model architecture we use to implement this framework.
% We first define some notation, then we will describe the framework in detail.

We begin with some notation.
Let $x$ be our source canvas and $y$ be our target canvas. In the regime of sequence modeling, a canvas is a sequence and we use the terms interchangeably.
%The source canvas would be the source sequence, and the target canvas would be the target sequence.
While this paper focuses on sequence generation, we note that our framework can be generalized to higher-dimensional outputs (e.g., image generation).

Let $\hat{y}_t$ be the hypothesis canvas at time $t$. Because our framework only supports insertions and not reordering operations, it must be a subsequence of the final output hypothesis $\hat{y}$. For example, if the eventual output were $\hat{y} = [A, B, C, D, E]$, then $\hat{y}_t = [B, D]$ would be a valid intermediate canvas while $\hat{y}_t = [B, A]$ would not. We do not restrict ourselves to one insertion per step, meaning $\hat{y}_t$ could have more than $t$ tokens.

% We define a hypothesis canvas $\hat{y_t}$ as a partially generated ordered subsequence of $y$ at iteration $t$. Note, $\hat{y_t}$ need not be generated left-to-right nor even have $t$ tokens. For example, in the case of $y = (A, B, C, D, E)$, $(B, D)$ would be a valid hypothesis of $y$, while $(B, A)$ would not be.

Further, let $\mathcal{C}$ be our content vocabulary (i.e., token vocabulary for sequences). At each iteration $t$, the Insertion Transformer produces a joint distribution over the choice of content $c \in \mathcal{C}$ and all available insertion locations $l \in [0, |\hat{y}_t|]$ in the current hypothesis canvas $\hat{y}_t$. In other words, the Insertion Transformer models both what to insert and where to insert relative to the current canvas hypothesis $\hat{y_t}$:
\begin{align}
    p(c, l \mid x, \hat{y_t}) = \mathrm{InsertionTransformer}(x, \hat{y_t}) .
\end{align}
As an example, suppose our current hypothesis canvas is $\hat{y}_t = [B, D]$ and we select the insertion operation $(c = C, l = 1)$.
%This says, let us insert content $c = E$ at location $l = 2$.
This will result in the new hypothesis canvas $\hat{y}_{t + 1} = [B, C, D]$. Also see Figure~\ref{fig:example-basic} for an example showing the full generation process for a typical English sentence.

The permitted insertion locations allow for insertions anywhere in the canvas from the leftmost slot ($l = 0$) to the rightmost slot ($l = |\hat{y}_t|$). Generation always begins with an empty canvas $\hat{y}_0 = []$ with just a single insertion location $l = 0$, and concludes when a special marker token is emitted. Exact details on termination handling can be found in Section \ref{subsec:termination}, where we describe two variants.

\section{Insertion Transformer Model}
\label{sec:model}

% of the one employed by
The concrete model we use for the Insertion Transformer is a modified version of the original Transformer \cite{vaswani-nips-2017}, with the decoder having been altered to induce a distribution over insertions anywhere in the current output rather than just at the end. We outline the key changes below.

%\subsection{Transformer Modifications}

%We list a few structural changes that are required for our model from the standard Transformer model \cite{vaswani-nips-2017}.

\paragraph{Full Decoder Self-Attention.}

We remove the causal self-attention mask from the decoder so that all positions can attend to all other positions, as opposed to just those to the left of the current position. This allows each decision to condition on the full context of the canvas hypothesis for the current iteration.

\paragraph{Slot Representations via Concatenated Outputs.}

The standard Transformer decoder produces $n$ vectors for a sequence of length $n$, one per position, with the last one being used to pick the next word. Our model instead requires $n + 1$ vectors, one for each of the $n - 1$ slots between words plus $2$ for the beginning and end slots. We achieve this by adding special marker tokens at the beginning and end of the decoder input to extend the sequence length by two. We then take the resulting $n + 2$ vectors in the final layer and concatenate each adjacent pair to obtain $n + 1$ slot representations. Hence each slot is summarized by the final representations of the positions to its immediate left and right.

\subsection{Model Variants}
\label{subsec:variants}

Beyond the required structural changes above, there are several variations of our model that we explore within our experiments.

\paragraph{Content-Location Distribution.}
We need to model the joint content-location distribution for the insertion operations. We present two approaches: the first directly models the joint distribution, the second relies on a factorization.

% We can model the joint distribution directly.
Let $H \in \mathbb{R}^{(T + 1) \times h}$ be the matrix of slot representations, where $h$ is the size of the hidden state and $T$ is the length of the current partial hypothesis. Let $W \in \mathbb{R}^{h \times |\mathcal{C}|}$ be the standard softmax projection matrix from the Transformer model. We can simply use this projection matrix to compute the content-location logits, then flatten this matrix into a vector and directly take the softmax over all the content-location logits to obtain a jointly normalized distribution:
\begin{align}
    p(c, l) = \mathrm{softmax}(\mathrm{flatten}(HW)) .
\end{align}

Another approach is to model the joint distribution using a conditional factorization, $p(c, l) = p(c \mid l) p(l)$.
We can model the conditional content distribution as is done in the normal Transformer:
\begin{align}
    p(c \mid l) = \mathrm{softmax}(h_l W) ,
\end{align}
where $h_l \in \mathbb{R}^{h}$ is the $l$-th row of $H$. In other words, we apply the softmax per-row in the matrix $HW$.
We separately model the location distribution by taking the softmax of the dot product of the hidden states and a learnable query vector $q \in \mathbb{R}^{h}$:
\begin{align}
    p(l) = \mathrm{softmax}(H q) .
\end{align}
This approach requires a small number of additional parameters $h$ compared to modeling the joint distribution directly.

% \paragraph{Joint Normalization over Content and Location.}

% Rather than decomposing the joint distribution over content $c$ and location $l$ as $p(c, l) = p(l) p(c, l)$, we can exclude the slot distribution $p(l)$ and take a single softmax over all the logits arising from the slot representations to directly obtain a full jointly normalized distribution.

% \paragraph{Slot Distribution via Attention.} For the variant of our approach that requires it, we model the distribution over slots by taking the dot product of a learned query vector with each slot representation, then computing the softmax of the resulting scores to obtain a probability distribution.

\paragraph{Contextualized Vocabulary Bias.}
To increase information sharing across slots, we can perform a max pooling operation over the final decoder hidden vectors $H$ to obtain a context vector $g \in \mathbb{R}^{h}$. We then project $g$ into the vocabulary space using a learned projection matrix $V \in \mathbb{R}^{h \times |\mathcal{C}|}$ to produce a shared bias $b \in \mathbb{R}^{|\mathcal{C}|}$. We then add $b$ to the result to the vocabulary logits at each position as an additional shared bias. We believe this may be useful in providing the model with coverage information, or in propagating count information about common words that should appear in multiple places in the output. Formally, we have
% problem below... need to add repmat or expand_dims equiv to `b`.
\begin{align}
    g &= \mathrm{maxpool}(H) \\
    b &= g V \\
    B &= \mathrm{repmat}(b, [T + 1, 1]) \\
    p(c, l) &= \mathrm{softmax}(HW + B)
\end{align}

\paragraph{Mixture-of-Softmaxes Output Layer.}

Unlike the output vectors of a typical autoregressive model which only need to capture distributional information about the next word, the slot vectors in our model are responsible for representing entire bags of words. Moreover, depending on the order of generation, they might correspond to any contiguous span of the final output, making this a highly nontrivial modeling problem. We posit that the language modeling softmax bottleneck identified by \citet{yang-iclr-2018} poses even greater challenges for our setup. We try including the mixture-of-softmaxes layer proposed in their work as one means of addressing the issue.

\section{Training and Loss Functions}

The Insertion Transformer framework is flexible enough to accommodate arbitrary generation orders, including those which are input- and context-dependent. We discuss several order loss functions that we can optimize for. % the Insertion Transformer.
% We can train Insertion Transformer to follow an expert order (i.e., left-to-right or balanced binary tree order), or we can train the Insertion Transformer to learn all possible orderings.

% We discuss two such possibilities in this section, then further describe a more neutral option in which entropy is maximized over the set of correct actions.

%Before diving into the details, however, we first must address
% In addition, we also address some key differences that arise when training an Insertion Transformer as compared with a standard autoregressive model such as the Transformer.

\subsection{Left-to-Right}

As a special case, the Insertion Transformer can be trained to produce its output in a left-to-right fashion, imitating the conventional setting where this ordering is enforced by construction.

To do so, given a training example $(x, y)$, we randomly sample a length $k \sim \text{Uniform}([0, |y|])$ and take the current hypothesis to be the left prefix $\hat{y} = (y_1, \dots, y_k)$. We then aim to maximize the probability of the next content in the sequence $c = y_k$ in the rightmost slot location $l = k$, using the negative log-likelihood of this action as our loss to be minimized:
\begin{align}
    \text{loss}(x, \hat{y}) = -\log p(y_{k+1}, k \mid x, \hat{y}) .
\end{align}
When the sequence is complete, i.e.\ $k = n$, we take $y_{k+1}$ to be the end-of-sequence token $\eos$.
% Not sure if we want to move this out of left-to-right, since the batch size problem is across all our losses.
We note that there are several differences between our left-to-right order loss and a standard autoregressive Transformer log-probability loss. We describe them in detail in Section \ref{sec:trainingdifferences}.

\subsection{Balanced Binary Tree}

% roll in policy: uniform
% target policy: exponentiated reward of distance from optimal target
% talk about temperature, parallel decoding, log n

% maybe use phrase recursively.
A left-to-right strategy only allows for one token to be inserted at a time. On the other end of the spectrum, we can train for maximal parallelism by using a balanced binary tree ordering. The centermost token is produced first, then the center tokens of the spans on either side are produced next, and this process is recursively continued until the full sequence has been generated. As an example, for the target output $[A, B, C, D, E, F, G]$, the desired order of production would be $[] \to [D] \to [B, D, F] \to [A, B, C, D, E, F, G]$, where multiple insertions are executed in parallel. See Section~\ref{sec:inference} for more details on parallel decoding.

To achieve this goal, we use a soft binary tree loss encouraging the model to assign high probability to tokens near the middle of the span represented by a given slot.
%and take the overall loss to be the average of these slot losses.
Partial canvas hypotheses are generated randomly so as to improve robustness and reduce exposure bias.

In more detail, given a training example $(x, y)$, we first sample a subsequence $\hat{y}$ from the set of all subsequences of the target $y$. One option would be to sample uniformly from this set, which could be accomplished by iterating through each token and keeping or throwing it out with probability $1/2$. Though simple, this approach would overexpose the model to partial outputs with length close to $|y|/2$ and would underexpose it to hypotheses that are nearly empty or nearly complete.

To circumvent this issue, we instead use a biased sampling procedure that gives uniform treatment to all lengths. In particular, we first sample a random length $k \sim \text{Uniform}([0, |y|])$, then sample a random subsequence of $y$ of length $k$. The latter step is carried out by constructing an index list $[1, \dots, |y|]$, shuffling it, and extracting the tokens corresponding to the first $k$ indices in the order they appear in the target sequence $y$.

\begin{figure}[t]
\centering
\includegraphics[width=\columnwidth]{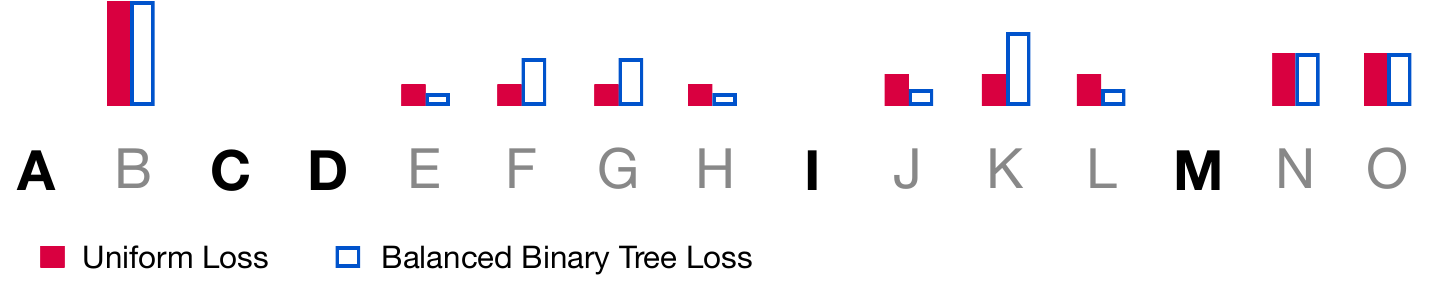}
\vspace{-2em}
\caption{A visualization of the weighting of the per-token negative log-likelihoods in the balanced binary tree and uniform losses. The balanced binary tree loss strongly incentivizes the generation of the center word or center words within each slot.}
\label{fig:losses}
\end{figure}

Once we have our randomly chosen hypothesis $\hat{y}$, it remains to compute the loss itself. For each of the $k + 1$ slots at locations $l = 0, \dots, k$, let $(y_{i_l}, y_{i_l + 1}, \dots, y_{j_l})$ be the span of tokens from the target output yet to be produced at location $l$. We first define a function $d_l$ giving the distance from the center of the span corresponding to location $l$:
\begin{align}
    d_l(i) &=  \left| \frac{i_l + j_l}{2} - i \right| .
\end{align}
We use the negative distance function $-d_l$ as the reward function for a softmax weighting policy $w_l$ \cite{rusu-iclr-2016,norouzi-nips-2016} (see Figure~\ref{fig:losses} for an illustration):
\begin{align}
    w_l(i) &= \frac{\exp(-d_l(i) / \tau)}{\sum_{i' = i_l}^{j_l} \exp(-d_l(i') / \tau)} .
\end{align}

Next we define the slot loss at location $l$ as a weighted sum of the negative log-likelihoods of the tokens from its corresponding span:
\begin{align}
    \text{slot-loss}(x, \hat{y}, l) &= \sum_{i = i_l}^{j_l} -\log p(y_i, l \mid x, \hat{y}) \cdot w_l(i) .
\end{align}
%
%, prioritizing those closest to the center using a distance-from-center function $d_l$
In other words, the loss encourages the model to prioritize the tokens closest to the center based on $d_l$. The temperature hyperparameter $\tau$ allows us to control the sharpness of the weight distribution, with $\tau \to 0$ approaching a peaked distribution placing all the weight on the centermost token (or centermost two tokens in the case of an even-length span), and $\tau \to \infty$ approaching a uniform distribution over all the missing content for a slot.

Finally, we define the full loss as the average of slot losses across all locations:
\begin{align}
    \text{loss}(x, \hat{y}) = \frac{1}{k+1} \sum_{l = 0}^k \text{slot-loss}(x, \hat{y}, l) .
\end{align}

\subsection{Uniform}

In addition to encouraging the model to follow a particular generation order, we can also train it to learn an agnostic view of the world in which it assigns equal probability mass to each correct action with no special preference. This neutral approach is useful insofar as it forces the model to be aware of all valid actions during each step of decoding, providing a rich learning signal during training and maximizing robustness.

Such an approach also bears resemblance to the principle of maximum entropy, which has successfully been employed for maximum entropy modeling across a number of domains in machine learning.
% (TODO: Add some citations here.)

To implement this loss, we simply take $\tau \to \infty$ in the binary tree loss of the previous section, yielding a slot loss of
\begin{align}
    \text{slot-loss}(x, \hat{y}, l) = \frac{1}{j_l - i_l + 1} \sum_{i = i_l}^{j_l} -\log p(y_i, l \mid x, \hat{y}) .
\end{align}
This is the mean of the negative log-probabilities of the correct actions for the given slot, which we note is maximized by a uniform distribution. Then as before, we take the full loss to be the mean of the slot losses.

\subsection{Termination}
\label{subsec:termination}

% There are a few possibilities for dealing with termination conditions when using the binary tree loss or the uniform loss.
We experiment with two termination conditions for the binary tree and uniform losses, slot finalization and sequence finalization, and compare their empirical performance in our experiments.

For slot finalization, when computing the slot loss for a location corresponding to an empty span in the true output, we take the target to be a single end-of-slot token. Then, all slot losses are always well-defined, and at generation time we can cease decoding when all slots predict an end-of-slot. We note for clarity that this special token appears in the vocabulary of the model but is never actually produced; see Section \ref{sec:inference} for more details.

Alternatively, for sequence finalization, we leave the slot losses undefined for empty spans and exclude them from the overall loss. Once the entire sequence has been produced and all locations correspond to empty spans, we take the slot loss at every location to be the negative log-likelihood of an end-of-sequence token. This is identical to the slot finalization approach at the very end, but differs while generation is ongoing as no signal is provided for empty slots.

\subsection{Training Differences}
\label{sec:trainingdifferences}
% We have access to the entire canvas $\hat{y}$, and therefore need to recompute the decoder states at every iteration. Consequently, there is no state (or gradient) propagation between iterations. Next, since we need to recompute the decoder states, we need to use an effective batch size an order of magnitude smaller than a regular Transformer due to memory limitations. In each batch sample, we can only optimize for 1 conditional, while in the case of Transformer, they can optimize for the entire sequence of conditionals.

In a typical neural autoregressive model, there is a unidirectional flow of information in the decoder. This allows hidden states to be propagated (and reused) across time steps during the generation process, since they will remain unaltered as the hypothesis is extended rightward. In contrast, because we allow for insertions anywhere in the sequence, our approach lacks this unidirectional property and we must recompute the decoder hidden states for each position after every insertion.

This has several consequences. First, there is no state (or gradient) propagation between generation steps. Next, instead of being able to efficiently compute the losses for all generation steps of an example in one fell swoop as is usually done, we can only compute the loss for one generation step at a time under the same memory constraints. Accordingly, our batch size is effectively reduced by a factor of the average sequence length, which has the potential to affect convergence speed and/or model quality. Finally, since we need to subsample generation steps during training, as opposed to a standard Transformer that can compute all the generation steps in a sequence for free, our gradient suffers from extra variance due to the sampling process. Under the right training conditions, however, we find these not to be major hindrances.
% Moreover, should we choose to subsample generation steps during training to maintain diversity within each batch, we lose the gradient propagation between generation steps that normally comes for free when training an autoregressive model.

% Hence there is a trade-off between model flexibility and ease of training, with the cost of our more general framework being some potential inefficiencies in learning. We explore this trade-off in more detail in the experimental section.

\section{Inference}
\label{sec:inference}

Recall that at each time step $t$, the Insertion Transformer yields a distribution $p(c, l \mid x, \hat{y}_t)$ over content $c$ and location $l$ given the input sequence $x$ and current partial output sequence $\hat{y}_t$. This highly flexible model opens the door for both sequential and parallel inference techniques, which we describe in more detail below.

\subsection{Greedy Decoding}

First we have a standard greedy approach to decoding, in which the action with the highest probability across all choices of content $c$ and location $l$ is selected:
\begin{align}
    (\hat{c}_t, \hat{l}_t) = \argmax_{c, l} p(c, l \mid x, \hat{y}_t) .
\end{align}

Once the best decision has been identified, we insert token $\hat{c}_t$ at location $\hat{l}_t$ to obtain the next partial output $\hat{y}_{t+1}$.

For models trained towards sequence finalization, this process continues until an end-of-sequence token gets selected at any location, at which point the final output is returned.

For models trained towards slot finalization, we restrict the $\argmax$ to locations whose maximum-probability decision is not end-of-slot, and finish only when the model predicts an end-of-slot token for every location.

\subsection{Parallel Decoding}

If we train an Insertion Transformer towards slot finalization, we can also parallelize inference across slots within each time step to obtain a simple partially autoregressive decoding algorithm.

In more detail, for each location $l$ we first compute the following maximum-probability actions:
\begin{align}
    \hat{c}_{l, t} = \argmax_{c} p(c \mid l, x, \hat{y}_t) .
\end{align}

For the version of the model whose joint distribution factors as $p(c, l) = p(l) p(c \mid l)$, the required conditional distribution $p(c \mid l)$ is already available. For the jointly normalized model, we can either obtain the conditional via renormalization as $p(c \mid l) = p(c, l) / p(l) = p(c, l) / \sum_{c'} p(c', l)$, or compute it directly by taking a softmax over the subset of logits at location $l$. In both cases, all the required conditional distributions can be computed in parallel.

Next, we filter out the locations for which the maximum-probability decision is an end-of-slot token, and for each location $l$ that remains, insert the selected token $\hat{c}_{l, t}$ into that slot. The resulting sequence becomes the next partial output $\hat{y}_{t+1}$. This process continues until an end-of-slot token is predicted at every location.

%%%% This is experimental section table -- here to move it to top of page.
\begin{table*}[t]
\centering
\small
%\resizebox{\columnwidth}{!}{%
\begin{tabular}{lcccc}
\toprule
\bfseries Loss & \bfseries Termination & \bfseries BLEU (+EOS) & \bfseries BLEU (+EOS) & \bfseries BLEU (+EOS) \\
\midrule
& & & \bfseries +Distillation & \bfseries +Distillation, +Parallel \\
%& & & & \bfseries  \\
\midrule
Left-to-Right & Sequence & 20.92 (20.92) & 23.29 (23.36) & - \\
\midrule
Binary Tree ($\tau = 0.5$) & Slot & 20.35 (21.39) & 24.49 (25.55) & 25.33 (25.70)\\
Binary Tree ($\tau = 1.0$) & Slot & 21.02 (22.37) & 24.36 (25.43) & 25.43 (25.76) \\
Binary Tree ($\tau = 2.0$) & Slot & 20.52 (21.95) & 24.59 (25.80) & 25.33 (25.80) \\
\midrule
Uniform & Sequence & 19.34 (22.64) & 22.75 (25.45) & - \\
Uniform & Slot & 18.26 (22.16) & 22.39 (25.58) & 24.31 (24.91) \\
\bottomrule
\end{tabular}
%}
\caption{Development BLEU scores obtained via greedy decoding for our basic models trained with various loss functions and termination strategies. The +EOS numbers are the BLEU score obtained when an EOS penalty is applied during decoding to discourage premature stopping. The +Distillation numbers are for models trained with distilled data. The +Parallel numbers are obtained with parallel decoding, which is applicable to models trained with the slot finalization termination condition.}
\label{tab:results-baseline}
\end{table*}

Since the parallel decoding scheme described here allows for a token to be inserted in every slot at every time step, a sequence of length $n$ could theoretically be generated in as few as $\lfloor \log_2 n \rfloor + 1$ steps. We find that this logarithmic complexity is attainable in practice in our experiments.

%%% TABLE IS IN paper_method to move the table to top of page.

\section{Experiments}

In this section, we explore the efficacy of our approach on a real-world machine translation task, analyzing its performance under different training conditions, architectural choices, and decoding procedures.
We experiment on the WMT 2014 English-German translation dataset, using newstest2013 for development and newstest2014 for testing, respectively. All our experiments are implemented in TensorFlow \cite{tensorflow-whitepaper-2015} using the Tensor2Tensor framework \cite{tensor2tensor}. We use the default \texttt{transformer\_base} hyperparameter set reported by \citet{tensor2tensor} for all hyperparameters not specific to our model. We perform no additional hyperparameter tuning. All our models are trained for 1,000,000 steps on eight P100 GPUs.

\subsection{Baseline Results}

We first train the baseline version of our model with different choices of loss functions and termination strategies. Greedy decoding results on the development set are given for each setting in the third column of Table~\ref{tab:results-baseline}.

We observe that the binary tree loss performs the best when standard greedy decoding is used, attaining a development BLEU score of 21.02. We also find that our left-to-right models do poorly compared to other orderings. One explanation is that the gradients of the binary tree and uniform losses are much more informative, in that they capture information on all the missing tokens, whereas left-to-right only provides information about the next one. We note that in all cases, even after 1M steps the models are still improving and do not appear to overfit.

Upon inspecting the outputs of these models, we found that some of the most common and severe mistakes were due to the model assigning high probability to the terminal token (end-of-slot or end-of-sequence, both abbreviated as EOS) too early in the decoding process, resulting in artificially short outputs.
To rectify this, we introduce an \textbf{EOS penalty} hyperparameter, which is a scalar subtracted from the log-probability assigned by the model to an EOS at each location during decoding. Using a penalty of $\beta$ prevents the model from selecting an EOS unless there is a difference of at least $\beta$ between the log-probability of EOS and the log-probability of the second-best choice. This approach is similar the length normalization techniques used in many sequence models \cite{graves-icml-2012}.
We perform a sweep over the range $[0, 7]$ and report the best result for each model in parentheses.
A well-chosen EOS penalty can have a sizable effect, increasing the BLEU score by nearly 4 points in some cases, and its inclusion brings the highest development score to 22.64 for the uniform loss with sequence-level finalization.

\subsection{Knowledge Distillation}

One technique shown to improve model performance on a wide variety of tasks is knowledge distillation \cite{hinton-nips-2015,kim-emnlp-2016}, wherein a model is trained on the outputs of another model.
%, possibly of a similar capacity and model class.
%We believe this practice could be especially useful for our approach, as it has the potential to smooth out noise in the data and reduce unnecessary variety, leading to an easier modeling problem. To make use of distillation, we first train a standard Transformer model on the same training corpus, then compute its predictions for each of the training examples. We use beam search with the hyperparameters from \citet{vaswani-nips-2017} to obtain the predictions. Following this, we train our proposed model on the distilled training data and rerun a subset of the baseline experiments from the previous section. The results are given in Table~\ref{tab:results-baseline}. 
We use the base Transformer model from \citet{vaswani-nips-2017} with beam search as our teacher model, and rerun a subset of the baseline experiments from the previous section on the resulting distilled data. The results are given in the fourth column of Table~\ref{tab:results-baseline}.

We observe improvements of 3 to 4 BLEU points across the board, showing that distillation is remarkably effective for our setting. As before, the models trained with a binary tree loss are approximately 2 BLEU points better than those trained with a uniform loss when standard decoding is performed, but the differences largely vanish when using a properly-tuned EOS penalty for each model. The best model by a small margin is the one trained with a binary tree loss with temperature $\tau = 2.0$, which achieves a 25.80 BLEU score on the development set.

\subsection{Architectural Variants}

Next we explore different combinations of the architectural variants described in Section~\ref{subsec:variants}. Using the uniform loss, slot finalization, and distillation as a neutral baseline configuration, we train each variant and decode on the development set to obtain the results given in Table~\ref{tab:results-variants}.

\begin{table}[t]
\centering
\small
%\resizebox{\columnwidth}{!}{%
\begin{tabular}{cccc}
\toprule
\bfseries Joint & \bfseries Contextual & \bfseries Mixture & \bfseries BLEU (+EOS) \\
\midrule
\textcolor{lightgray}{\xmark} & \textcolor{lightgray}{\xmark} & \textcolor{lightgray}{\xmark} & 22.39 (25.58) \\
\cmark & \textcolor{lightgray}{\xmark} & \textcolor{lightgray}{\xmark} & 22.92 (25.14) \\
\textcolor{lightgray}{\xmark} & \cmark & \textcolor{lightgray}{\xmark} & 23.00 (25.41) \\
\textcolor{lightgray}{\xmark} &\textcolor{lightgray}{\xmark} & \cmark & 22.19 (25.58) \\
\cmark & \cmark & \textcolor{lightgray}{\xmark} & 23.22 (25.44) \\
\cmark & \textcolor{lightgray}{\xmark} & \cmark & 20.17 (24.19) \\
\textcolor{lightgray}{\xmark} & \cmark & \cmark & 23.29 (25.48) \\
\cmark & \cmark & \cmark & 22.16 (25.44) \\
\bottomrule
\end{tabular}
%}
\caption{Development BLEU scores obtained via greedy decoding when training models with the architectural variants discussed in Section~\ref{subsec:variants}. All models are trained with a uniform loss and slot finalization on distilled data.}
\label{tab:results-variants}
\end{table}

Many of the configurations help improve performance when decoding without an EOS penalty. In particular, using joint normalization, a contextualized vocabulary bias, or both leads to improvements of 0.5-0.8 BLEU over the baseline. Once we tune the EOS penalty for each setting, however, the improvements largely disappear. The best configurations, primarily those involving mixture-of-softmaxes, are within 0.1 BLEU of the baseline. This suggests that the core architecture is already sufficiently powerful when decoding is well-tuned, but that it may be useful to consider some variations when looking at other inference settings.

\subsection{Parallel Decoding}

Thus far, all our experiments have used greedy decoding. However, as described in Section~\ref{sec:inference}, models trained towards slot finalization also permit a parallel decoding scheme in which tokens are simultaneously inserted into every unfinished slot at each time step until no such slots remain. We decode the development set using this strategy for some of our more promising models, giving results in Table~\ref{tab:results-parallel}. Some example decodes are provided in Figure~\ref{fig:example-parallel} for reference.

\begin{table}[t]
\centering
\small
%\resizebox{\columnwidth}{!}{%
\begin{tabular}{lc}
\toprule
\bfseries Model & \bfseries BLEU (+EOS) \\
\midrule
Binary Tree ($\tau = 0.5$) & 25.33 (25.70) \\
Binary Tree ($\tau = 1.0$) & 25.43 (25.76) \\
Binary Tree ($\tau = 2.0$) & 25.33 (25.80) \\
\midrule
Uniform & 24.31 (24.91) \\
Uniform + Contextual & 24.54 (24.74) \\
Uniform + Mixture & 24.33 (25.11) \\
Uniform + Contextual + Mixture & 24.68 (25.02) \\
\bottomrule
\end{tabular}
%}
\caption{Parallel decoding results on the development set for some of our stronger models. All numbers are comparable to or even slightly better than those obtained via greedy decoding, demonstrating that our model can perform insertions in parallel with little to no cost for end performance.}
\label{tab:results-parallel}
\end{table}

First and foremost, we observe that all scores are on par with those obtained via greedy decoding, and in some cases are even better. This demonstrates that with a proper training objective, our model can seamlessly accommodate parallel insertions with little effect on end performance. The fact that some scores are improved suggests that greedy search may suffer from issues related to local search that are circumvented by making multiple updates to the hypothesis at once. We leave this as an interesting topic for future investigation.

In addition, we find that parallel decoding also helps close the gap between results obtained with and without an EOS penalty. We believe this may be due in part to the fact that the number of decoding iterations is reduced substantially, thereby giving fewer opportunities for the model to erroneously stop at an intermediate state.

\begin{figure}[t]
\centering
\pgfplotstableread{parallel-decoding-steps-btree.txt}{\plotDataBtree}
\pgfplotstableread{parallel-decoding-steps-uniform.txt}{\plotDataUniform}
\begin{tikzpicture}
\begin{axis}[
  xlabel={Output Length},
  ylabel={Number of Decoding Iterations},
  legend style={at={(0.04,0.96)},anchor=north west,font=\small},
  legend cell align={left},
  height=0.7\columnwidth,
  width=\columnwidth,
]
\addplot[mark=none,black] table[x=length,y=length] from \plotDataBtree;
\addplot[mark=none,red] table[x=length,y=min_steps] from \plotDataBtree;
\addplot[only marks,mark=x,green!70!black] table[x=length,y=num_steps] from \plotDataUniform; 
\addplot[only marks,mark=+,blue] table[x=length,y=num_steps] from \plotDataBtree;
\legend{Upper Bound ($n$),Lower Bound ($\lfloor \log_2 n \rfloor + 1$),Insertion Transformer (Uniform),Insertion Transformer (Binary Tree)}
\end{axis}
\end{tikzpicture}
\vspace{-1em}
\caption{Plot showing number of decoding iterations versus output length as measured on the development set for our best models. To produce an output of length $n$, an insertion-based model requires at least $\lfloor \log_2 n \rfloor + 1$ iterations and at most $n$. While greedy decoding cannot do better than the upper bound, our parallel decoding scheme nearly achieves the lower bound in all cases.}
\label{fig:plot-parallel}
\end{figure}
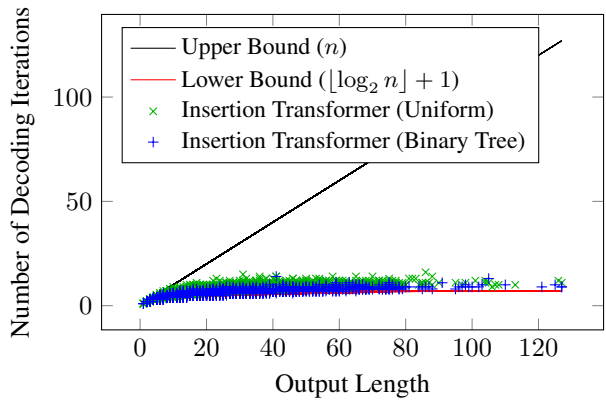

\begin{figure*}[t]
\small
\begin{flushleft}
\textbf{Input:} But on the other side of the state, that is not the impression many people have of their former governor. \\
\vspace{0.5em}
\textbf{Output:} Aber auf der anderen Seite des Staates ist das nicht der Eindruck, den viele von ihrem ehemaligen Gouverneur haben. \\
\vspace{0.5em}
\textbf{Parallel decode (binary tree loss):}
\end{flushleft}
\vspace{-1.5em}
\begin{center}
\resizebox{\textwidth}{!}{%
\begin{tikzpicture}[every node/.style={anchor=base,inner sep=0.1em}]
\node[gray!60] (node-0-0) {Aber\_};
\node[gray!60,base right=0cm of node-0-0] (node-0-1) {auf\_};
\node[gray!60,base right=0cm of node-0-1] (node-0-2) {der\_};
\node[gray!60,base right=0cm of node-0-2] (node-0-3) {anderen\_};
\node[gray!60,base right=0cm of node-0-3] (node-0-4) {Seite\_};
\node[gray!60,base right=0cm of node-0-4] (node-0-5) {des\_};
\node[gray!60,base right=0cm of node-0-5] (node-0-6) {Staates\_};
\node[gray!60,base right=0cm of node-0-6] (node-0-7) {ist\_};
\node[gray!60,base right=0cm of node-0-7] (node-0-8) {das\_};
\node[gray!60,base right=0cm of node-0-8] (node-0-9) {nicht\_};
\node[gray!60,base right=0cm of node-0-9] (node-0-10) {der\_};
\node[blue,base right=0cm of node-0-10] (node-0-11) {\uline{Eindruck\_}};
\node[gray!60,base right=0cm of node-0-11] (node-0-12) {, \_};
\node[gray!60,base right=0cm of node-0-12] (node-0-13) {den\_};
\node[gray!60,base right=0cm of node-0-13] (node-0-14) {viele\_};
\node[gray!60,base right=0cm of node-0-14] (node-0-15) {von\_};
\node[gray!60,base right=0cm of node-0-15] (node-0-16) {ihrem\_};
\node[gray!60,base right=0cm of node-0-16] (node-0-17) {ehemaligen\_};
\node[gray!60,base right=0cm of node-0-17] (node-0-18) {Gouverneur};
\node[gray!60,base right=0cm of node-0-18] (node-0-19) {\_};
\node[gray!60,base right=0cm of node-0-19] (node-0-20) {haben\_};
\node[gray!60,base right=0cm of node-0-20] (node-0-21) {.\_};
\node[gray!60,below=0.1cm of node-0-0] (node-1-0) {Aber\_};
\node[gray!60,base right=0cm of node-1-0] (node-1-1) {auf\_};
\node[gray!60,base right=0cm of node-1-1] (node-1-2) {der\_};
\node[gray!60,base right=0cm of node-1-2] (node-1-3) {anderen\_};
\node[gray!60,base right=0cm of node-1-3] (node-1-4) {Seite\_};
\node[blue,base right=0cm of node-1-4] (node-1-5) {\uline{des\_}};
\node[gray!60,base right=0cm of node-1-5] (node-1-6) {Staates\_};
\node[gray!60,base right=0cm of node-1-6] (node-1-7) {ist\_};
\node[gray!60,base right=0cm of node-1-7] (node-1-8) {das\_};
\node[gray!60,base right=0cm of node-1-8] (node-1-9) {nicht\_};
\node[gray!60,base right=0cm of node-1-9] (node-1-10) {der\_};
\node[black,base right=0cm of node-1-10] (node-1-11) {Eindruck\_};
\node[gray!60,base right=0cm of node-1-11] (node-1-12) {, \_};
\node[gray!60,base right=0cm of node-1-12] (node-1-13) {den\_};
\node[gray!60,base right=0cm of node-1-13] (node-1-14) {viele\_};
\node[gray!60,base right=0cm of node-1-14] (node-1-15) {von\_};
\node[gray!60,base right=0cm of node-1-15] (node-1-16) {ihrem\_};
\node[blue,base right=0cm of node-1-16] (node-1-17) {\uline{ehemaligen\_}};
\node[gray!60,base right=0cm of node-1-17] (node-1-18) {Gouverneur};
\node[gray!60,base right=0cm of node-1-18] (node-1-19) {\_};
\node[gray!60,base right=0cm of node-1-19] (node-1-20) {haben\_};
\node[gray!60,base right=0cm of node-1-20] (node-1-21) {.\_};
\node[gray!60,below=0.1cm of node-1-0] (node-2-0) {Aber\_};
\node[gray!60,base right=0cm of node-2-0] (node-2-1) {auf\_};
\node[blue,base right=0cm of node-2-1] (node-2-2) {\uline{der\_}};
\node[gray!60,base right=0cm of node-2-2] (node-2-3) {anderen\_};
\node[gray!60,base right=0cm of node-2-3] (node-2-4) {Seite\_};
\node[black,base right=0cm of node-2-4] (node-2-5) {des\_};
\node[gray!60,base right=0cm of node-2-5] (node-2-6) {Staates\_};
\node[gray!60,base right=0cm of node-2-6] (node-2-7) {ist\_};
\node[gray!60,base right=0cm of node-2-7] (node-2-8) {das\_};
\node[blue,base right=0cm of node-2-8] (node-2-9) {\uline{nicht\_}};
\node[gray!60,base right=0cm of node-2-9] (node-2-10) {der\_};
\node[black,base right=0cm of node-2-10] (node-2-11) {Eindruck\_};
\node[gray!60,base right=0cm of node-2-11] (node-2-12) {, \_};
\node[gray!60,base right=0cm of node-2-12] (node-2-13) {den\_};
\node[blue,base right=0cm of node-2-13] (node-2-14) {\uline{viele\_}};
\node[gray!60,base right=0cm of node-2-14] (node-2-15) {von\_};
\node[gray!60,base right=0cm of node-2-15] (node-2-16) {ihrem\_};
\node[black,base right=0cm of node-2-16] (node-2-17) {ehemaligen\_};
\node[gray!60,base right=0cm of node-2-17] (node-2-18) {Gouverneur};
\node[gray!60,base right=0cm of node-2-18] (node-2-19) {\_};
\node[blue,base right=0cm of node-2-19] (node-2-20) {\uline{haben\_}};
\node[gray!60,base right=0cm of node-2-20] (node-2-21) {.\_};
\node[blue,below=0.1cm of node-2-0] (node-3-0) {\uline{Aber\_}};
\node[gray!60,base right=0cm of node-3-0] (node-3-1) {auf\_};
\node[black,base right=0cm of node-3-1] (node-3-2) {der\_};
\node[blue,base right=0cm of node-3-2] (node-3-3) {\uline{anderen\_}};
\node[gray!60,base right=0cm of node-3-3] (node-3-4) {Seite\_};
\node[black,base right=0cm of node-3-4] (node-3-5) {des\_};
\node[gray!60,base right=0cm of node-3-5] (node-3-6) {Staates\_};
\node[blue,base right=0cm of node-3-6] (node-3-7) {\uline{ist\_}};
\node[gray!60,base right=0cm of node-3-7] (node-3-8) {das\_};
\node[black,base right=0cm of node-3-8] (node-3-9) {nicht\_};
\node[blue,base right=0cm of node-3-9] (node-3-10) {\uline{der\_}};
\node[black,base right=0cm of node-3-10] (node-3-11) {Eindruck\_};
\node[gray!60,base right=0cm of node-3-11] (node-3-12) {, \_};
\node[blue,base right=0cm of node-3-12] (node-3-13) {\uline{den\_}};
\node[black,base right=0cm of node-3-13] (node-3-14) {viele\_};
\node[gray!60,base right=0cm of node-3-14] (node-3-15) {von\_};
\node[blue,base right=0cm of node-3-15] (node-3-16) {\uline{ihrem\_}};
\node[black,base right=0cm of node-3-16] (node-3-17) {ehemaligen\_};
\node[blue,base right=0cm of node-3-17] (node-3-18) {\uline{Gouverneur}};
\node[gray!60,base right=0cm of node-3-18] (node-3-19) {\_};
\node[black,base right=0cm of node-3-19] (node-3-20) {haben\_};
\node[blue,base right=0cm of node-3-20] (node-3-21) {\uline{.\_}};
\node[black,below=0.1cm of node-3-0] (node-4-0) {Aber\_};
\node[blue,base right=0cm of node-4-0] (node-4-1) {\uline{auf\_}};
\node[black,base right=0cm of node-4-1] (node-4-2) {der\_};
\node[black,base right=0cm of node-4-2] (node-4-3) {anderen\_};
\node[blue,base right=0cm of node-4-3] (node-4-4) {\uline{Seite\_}};
\node[black,base right=0cm of node-4-4] (node-4-5) {des\_};
\node[blue,base right=0cm of node-4-5] (node-4-6) {\uline{Staates\_}};
\node[black,base right=0cm of node-4-6] (node-4-7) {ist\_};
\node[blue,base right=0cm of node-4-7] (node-4-8) {\uline{das\_}};
\node[black,base right=0cm of node-4-8] (node-4-9) {nicht\_};
\node[black,base right=0cm of node-4-9] (node-4-10) {der\_};
\node[black,base right=0cm of node-4-10] (node-4-11) {Eindruck\_};
\node[blue,base right=0cm of node-4-11] (node-4-12) {\uline{, \_}};
\node[black,base right=0cm of node-4-12] (node-4-13) {den\_};
\node[black,base right=0cm of node-4-13] (node-4-14) {viele\_};
\node[blue,base right=0cm of node-4-14] (node-4-15) {\uline{von\_}};
\node[black,base right=0cm of node-4-15] (node-4-16) {ihrem\_};
\node[black,base right=0cm of node-4-16] (node-4-17) {ehemaligen\_};
\node[black,base right=0cm of node-4-17] (node-4-18) {Gouverneur};
\node[blue,base right=0cm of node-4-18] (node-4-19) {\uline{\_}};
\node[black,base right=0cm of node-4-19] (node-4-20) {haben\_};
\node[black,base right=0cm of node-4-20] (node-4-21) {.\_};
\end{tikzpicture}
}
\end{center}
\vspace{-0.5em}
\parbox{\textwidth}{\dotfill}
\begin{flushleft}
\textbf{Input:} Everyone has the Internet, an iPad and eBooks. \\
\vspace{0.5em}
\textbf{Output:} Jeder hat das Internet, ein iPad und eBooks. \\
\vspace{0.5em}
\textbf{Greedy decode (uniform loss):}
\end{flushleft}
\vspace{-1.2em}
\begin{flushleft}
\resizebox{0.8\textwidth}{!}{%
\begin{tikzpicture}[every node/.style={anchor=base,inner sep=0.1em},baseline=(current bounding box.north)]
\node[gray!60] (node-0-0) {Jeder\_};
\node[gray!60,base right=0cm of node-0-0] (node-0-1) {hat\_};
\node[gray!60,base right=0cm of node-0-1] (node-0-2) {das\_};
\node[gray!60,base right=0cm of node-0-2] (node-0-3) {Internet\_};
\node[gray!60,base right=0cm of node-0-3] (node-0-4) {, \_};
\node[gray!60,base right=0cm of node-0-4] (node-0-5) {ein\_};
\node[gray!60,base right=0cm of node-0-5] (node-0-6) {i};
\node[gray!60,base right=0cm of node-0-6] (node-0-7) {Pad};
\node[gray!60,base right=0cm of node-0-7] (node-0-8) {\_};
\node[gray!60,base right=0cm of node-0-8] (node-0-9) {und\_};
\node[gray!60,base right=0cm of node-0-9] (node-0-10) {eB};
\node[blue,base right=0cm of node-0-10] (node-0-11) {\uline{oo}};
\node[gray!60,base right=0cm of node-0-11] (node-0-12) {ks\_};
\node[gray!60,base right=0cm of node-0-12] (node-0-13) {.\_};
\node[gray!60,below=0.1cm of node-0-0] (node-1-0) {Jeder\_};
\node[gray!60,base right=0cm of node-1-0] (node-1-1) {hat\_};
\node[gray!60,base right=0cm of node-1-1] (node-1-2) {das\_};
\node[gray!60,base right=0cm of node-1-2] (node-1-3) {Internet\_};
\node[gray!60,base right=0cm of node-1-3] (node-1-4) {, \_};
\node[gray!60,base right=0cm of node-1-4] (node-1-5) {ein\_};
\node[gray!60,base right=0cm of node-1-5] (node-1-6) {i};
\node[blue,base right=0cm of node-1-6] (node-1-7) {\uline{Pad}};
\node[gray!60,base right=0cm of node-1-7] (node-1-8) {\_};
\node[gray!60,base right=0cm of node-1-8] (node-1-9) {und\_};
\node[gray!60,base right=0cm of node-1-9] (node-1-10) {eB};
\node[black,base right=0cm of node-1-10] (node-1-11) {oo};
\node[gray!60,base right=0cm of node-1-11] (node-1-12) {ks\_};
\node[gray!60,base right=0cm of node-1-12] (node-1-13) {.\_};
\node[blue,below=0.1cm of node-1-0] (node-2-0) {\uline{Jeder\_}};
\node[gray!60,base right=0cm of node-2-0] (node-2-1) {hat\_};
\node[gray!60,base right=0cm of node-2-1] (node-2-2) {das\_};
\node[gray!60,base right=0cm of node-2-2] (node-2-3) {Internet\_};
\node[gray!60,base right=0cm of node-2-3] (node-2-4) {, \_};
\node[gray!60,base right=0cm of node-2-4] (node-2-5) {ein\_};
\node[gray!60,base right=0cm of node-2-5] (node-2-6) {i};
\node[black,base right=0cm of node-2-6] (node-2-7) {Pad};
\node[gray!60,base right=0cm of node-2-7] (node-2-8) {\_};
\node[gray!60,base right=0cm of node-2-8] (node-2-9) {und\_};
\node[gray!60,base right=0cm of node-2-9] (node-2-10) {eB};
\node[black,base right=0cm of node-2-10] (node-2-11) {oo};
\node[gray!60,base right=0cm of node-2-11] (node-2-12) {ks\_};
\node[gray!60,base right=0cm of node-2-12] (node-2-13) {.\_};
\node[black,below=0.1cm of node-2-0] (node-3-0) {Jeder\_};
\node[gray!60,base right=0cm of node-3-0] (node-3-1) {hat\_};
\node[blue,base right=0cm of node-3-1] (node-3-2) {\uline{das\_}};
\node[gray!60,base right=0cm of node-3-2] (node-3-3) {Internet\_};
\node[gray!60,base right=0cm of node-3-3] (node-3-4) {, \_};
\node[gray!60,base right=0cm of node-3-4] (node-3-5) {ein\_};
\node[gray!60,base right=0cm of node-3-5] (node-3-6) {i};
\node[black,base right=0cm of node-3-6] (node-3-7) {Pad};
\node[gray!60,base right=0cm of node-3-7] (node-3-8) {\_};
\node[gray!60,base right=0cm of node-3-8] (node-3-9) {und\_};
\node[gray!60,base right=0cm of node-3-9] (node-3-10) {eB};
\node[black,base right=0cm of node-3-10] (node-3-11) {oo};
\node[gray!60,base right=0cm of node-3-11] (node-3-12) {ks\_};
\node[gray!60,base right=0cm of node-3-12] (node-3-13) {.\_};
\node[black,below=0.1cm of node-3-0] (node-4-0) {Jeder\_};
\node[gray!60,base right=0cm of node-4-0] (node-4-1) {hat\_};
\node[black,base right=0cm of node-4-1] (node-4-2) {das\_};
\node[gray!60,base right=0cm of node-4-2] (node-4-3) {Internet\_};
\node[blue,base right=0cm of node-4-3] (node-4-4) {\uline{, \_}};
\node[gray!60,base right=0cm of node-4-4] (node-4-5) {ein\_};
\node[gray!60,base right=0cm of node-4-5] (node-4-6) {i};
\node[black,base right=0cm of node-4-6] (node-4-7) {Pad};
\node[gray!60,base right=0cm of node-4-7] (node-4-8) {\_};
\node[gray!60,base right=0cm of node-4-8] (node-4-9) {und\_};
\node[gray!60,base right=0cm of node-4-9] (node-4-10) {eB};
\node[black,base right=0cm of node-4-10] (node-4-11) {oo};
\node[gray!60,base right=0cm of node-4-11] (node-4-12) {ks\_};
\node[gray!60,base right=0cm of node-4-12] (node-4-13) {.\_};
\node[black,below=0.1cm of node-4-0] (node-5-0) {Jeder\_};
\node[gray!60,base right=0cm of node-5-0] (node-5-1) {hat\_};
\node[black,base right=0cm of node-5-1] (node-5-2) {das\_};
\node[gray!60,base right=0cm of node-5-2] (node-5-3) {Internet\_};
\node[black,base right=0cm of node-5-3] (node-5-4) {, \_};
\node[gray!60,base right=0cm of node-5-4] (node-5-5) {ein\_};
\node[gray!60,base right=0cm of node-5-5] (node-5-6) {i};
\node[black,base right=0cm of node-5-6] (node-5-7) {Pad};
\node[gray!60,base right=0cm of node-5-7] (node-5-8) {\_};
\node[gray!60,base right=0cm of node-5-8] (node-5-9) {und\_};
\node[blue,base right=0cm of node-5-9] (node-5-10) {\uline{eB}};
\node[black,base right=0cm of node-5-10] (node-5-11) {oo};
\node[gray!60,base right=0cm of node-5-11] (node-5-12) {ks\_};
\node[gray!60,base right=0cm of node-5-12] (node-5-13) {.\_};
\node[black,below=0.1cm of node-5-0] (node-6-0) {Jeder\_};
\node[gray!60,base right=0cm of node-6-0] (node-6-1) {hat\_};
\node[black,base right=0cm of node-6-1] (node-6-2) {das\_};
\node[gray!60,base right=0cm of node-6-2] (node-6-3) {Internet\_};
\node[black,base right=0cm of node-6-3] (node-6-4) {, \_};
\node[gray!60,base right=0cm of node-6-4] (node-6-5) {ein\_};
\node[gray!60,base right=0cm of node-6-5] (node-6-6) {i};
\node[black,base right=0cm of node-6-6] (node-6-7) {Pad};
\node[gray!60,base right=0cm of node-6-7] (node-6-8) {\_};
\node[gray!60,base right=0cm of node-6-8] (node-6-9) {und\_};
\node[black,base right=0cm of node-6-9] (node-6-10) {eB};
\node[black,base right=0cm of node-6-10] (node-6-11) {oo};
\node[blue,base right=0cm of node-6-11] (node-6-12) {\uline{ks\_}};
\node[gray!60,base right=0cm of node-6-12] (node-6-13) {.\_};
\node[black,below=0.1cm of node-6-0] (node-7-0) {Jeder\_};
\node[gray!60,base right=0cm of node-7-0] (node-7-1) {hat\_};
\node[black,base right=0cm of node-7-1] (node-7-2) {das\_};
\node[gray!60,base right=0cm of node-7-2] (node-7-3) {Internet\_};
\node[black,base right=0cm of node-7-3] (node-7-4) {, \_};
\node[gray!60,base right=0cm of node-7-4] (node-7-5) {ein\_};
\node[gray!60,base right=0cm of node-7-5] (node-7-6) {i};
\node[black,base right=0cm of node-7-6] (node-7-7) {Pad};
\node[gray!60,base right=0cm of node-7-7] (node-7-8) {\_};
\node[blue,base right=0cm of node-7-8] (node-7-9) {\uline{und\_}};
\node[black,base right=0cm of node-7-9] (node-7-10) {eB};
\node[black,base right=0cm of node-7-10] (node-7-11) {oo};
\node[black,base right=0cm of node-7-11] (node-7-12) {ks\_};
\node[gray!60,base right=0cm of node-7-12] (node-7-13) {.\_};
\end{tikzpicture}
\hspace{3em}
\begin{tikzpicture}[every node/.style={anchor=base,inner sep=0.1em},baseline=(current bounding box.north)]
\node (node-7-0) {(continued)};
%\node[black,below=0.1cm of node-7-0] (node-8-0) {Jeder\_};
\node[black,below=0.5cm of node-7-0.base west,anchor=base west] (node-8-0) {Jeder\_};
%\node[black] (node-8-0) {Jeder\_};
\node[gray!60,base right=0cm of node-8-0] (node-8-1) {hat\_};
\node[black,base right=0cm of node-8-1] (node-8-2) {das\_};
\node[gray!60,base right=0cm of node-8-2] (node-8-3) {Internet\_};
\node[black,base right=0cm of node-8-3] (node-8-4) {, \_};
\node[gray!60,base right=0cm of node-8-4] (node-8-5) {ein\_};
\node[blue,base right=0cm of node-8-5] (node-8-6) {\uline{i}};
\node[black,base right=0cm of node-8-6] (node-8-7) {Pad};
\node[gray!60,base right=0cm of node-8-7] (node-8-8) {\_};
\node[black,base right=0cm of node-8-8] (node-8-9) {und\_};
\node[black,base right=0cm of node-8-9] (node-8-10) {eB};
\node[black,base right=0cm of node-8-10] (node-8-11) {oo};
\node[black,base right=0cm of node-8-11] (node-8-12) {ks\_};
\node[gray!60,base right=0cm of node-8-12] (node-8-13) {.\_};
\node[black,below=0.1cm of node-8-0] (node-9-0) {Jeder\_};
\node[gray!60,base right=0cm of node-9-0] (node-9-1) {hat\_};
\node[black,base right=0cm of node-9-1] (node-9-2) {das\_};
\node[gray!60,base right=0cm of node-9-2] (node-9-3) {Internet\_};
\node[black,base right=0cm of node-9-3] (node-9-4) {, \_};
\node[gray!60,base right=0cm of node-9-4] (node-9-5) {ein\_};
\node[black,base right=0cm of node-9-5] (node-9-6) {i};
\node[black,base right=0cm of node-9-6] (node-9-7) {Pad};
\node[gray!60,base right=0cm of node-9-7] (node-9-8) {\_};
\node[black,base right=0cm of node-9-8] (node-9-9) {und\_};
\node[black,base right=0cm of node-9-9] (node-9-10) {eB};
\node[black,base right=0cm of node-9-10] (node-9-11) {oo};
\node[black,base right=0cm of node-9-11] (node-9-12) {ks\_};
\node[blue,base right=0cm of node-9-12] (node-9-13) {\uline{.\_}};
\node[black,below=0.1cm of node-9-0] (node-10-0) {Jeder\_};
\node[gray!60,base right=0cm of node-10-0] (node-10-1) {hat\_};
\node[black,base right=0cm of node-10-1] (node-10-2) {das\_};
\node[gray!60,base right=0cm of node-10-2] (node-10-3) {Internet\_};
\node[black,base right=0cm of node-10-3] (node-10-4) {, \_};
\node[gray!60,base right=0cm of node-10-4] (node-10-5) {ein\_};
\node[black,base right=0cm of node-10-5] (node-10-6) {i};
\node[black,base right=0cm of node-10-6] (node-10-7) {Pad};
\node[blue,base right=0cm of node-10-7] (node-10-8) {\uline{\_}};
\node[black,base right=0cm of node-10-8] (node-10-9) {und\_};
\node[black,base right=0cm of node-10-9] (node-10-10) {eB};
\node[black,base right=0cm of node-10-10] (node-10-11) {oo};
\node[black,base right=0cm of node-10-11] (node-10-12) {ks\_};
\node[black,base right=0cm of node-10-12] (node-10-13) {.\_};
\node[black,below=0.1cm of node-10-0] (node-11-0) {Jeder\_};
\node[blue,base right=0cm of node-11-0] (node-11-1) {\uline{hat\_}};
\node[black,base right=0cm of node-11-1] (node-11-2) {das\_};
\node[gray!60,base right=0cm of node-11-2] (node-11-3) {Internet\_};
\node[black,base right=0cm of node-11-3] (node-11-4) {, \_};
\node[gray!60,base right=0cm of node-11-4] (node-11-5) {ein\_};
\node[black,base right=0cm of node-11-5] (node-11-6) {i};
\node[black,base right=0cm of node-11-6] (node-11-7) {Pad};
\node[black,base right=0cm of node-11-7] (node-11-8) {\_};
\node[black,base right=0cm of node-11-8] (node-11-9) {und\_};
\node[black,base right=0cm of node-11-9] (node-11-10) {eB};
\node[black,base right=0cm of node-11-10] (node-11-11) {oo};
\node[black,base right=0cm of node-11-11] (node-11-12) {ks\_};
\node[black,base right=0cm of node-11-12] (node-11-13) {.\_};
\node[black,below=0.1cm of node-11-0] (node-12-0) {Jeder\_};
\node[black,base right=0cm of node-12-0] (node-12-1) {hat\_};
\node[black,base right=0cm of node-12-1] (node-12-2) {das\_};
\node[blue,base right=0cm of node-12-2] (node-12-3) {\uline{Internet\_}};
\node[black,base right=0cm of node-12-3] (node-12-4) {, \_};
\node[gray!60,base right=0cm of node-12-4] (node-12-5) {ein\_};
\node[black,base right=0cm of node-12-5] (node-12-6) {i};
\node[black,base right=0cm of node-12-6] (node-12-7) {Pad};
\node[black,base right=0cm of node-12-7] (node-12-8) {\_};
\node[black,base right=0cm of node-12-8] (node-12-9) {und\_};
\node[black,base right=0cm of node-12-9] (node-12-10) {eB};
\node[black,base right=0cm of node-12-10] (node-12-11) {oo};
\node[black,base right=0cm of node-12-11] (node-12-12) {ks\_};
\node[black,base right=0cm of node-12-12] (node-12-13) {.\_};
\node[black,below=0.1cm of node-12-0] (node-13-0) {Jeder\_};
\node[black,base right=0cm of node-13-0] (node-13-1) {hat\_};
\node[black,base right=0cm of node-13-1] (node-13-2) {das\_};
\node[black,base right=0cm of node-13-2] (node-13-3) {Internet\_};
\node[black,base right=0cm of node-13-3] (node-13-4) {, \_};
\node[blue,base right=0cm of node-13-4] (node-13-5) {\uline{ein\_}};
\node[black,base right=0cm of node-13-5] (node-13-6) {i};
\node[black,base right=0cm of node-13-6] (node-13-7) {Pad};
\node[black,base right=0cm of node-13-7] (node-13-8) {\_};
\node[black,base right=0cm of node-13-8] (node-13-9) {und\_};
\node[black,base right=0cm of node-13-9] (node-13-10) {eB};
\node[black,base right=0cm of node-13-10] (node-13-11) {oo};
\node[black,base right=0cm of node-13-11] (node-13-12) {ks\_};
\node[black,base right=0cm of node-13-12] (node-13-13) {.\_};
\node[black,below=0.1cm of node-13-0] (node-14-0) {Jeder\_};
\node[black,base right=0cm of node-14-0] (node-14-1) {hat\_};
\node[black,base right=0cm of node-14-1] (node-14-2) {das\_};
\node[black,base right=0cm of node-14-2] (node-14-3) {Internet\_};
\node[black,base right=0cm of node-14-3] (node-14-4) {, \_};
\node[black,base right=0cm of node-14-4] (node-14-5) {ein\_};
\node[black,base right=0cm of node-14-5] (node-14-6) {i};
\node[black,base right=0cm of node-14-6] (node-14-7) {Pad};
\node[black,base right=0cm of node-14-7] (node-14-8) {\_};
\node[black,base right=0cm of node-14-8] (node-14-9) {und\_};
\node[black,base right=0cm of node-14-9] (node-14-10) {eB};
\node[black,base right=0cm of node-14-10] (node-14-11) {oo};
\node[black,base right=0cm of node-14-11] (node-14-12) {ks\_};
\node[black,base right=0cm of node-14-12] (node-14-13) {.\_};
\end{tikzpicture}
}
\end{flushleft}
\parbox{\textwidth}{\dotfill}
\begin{flushleft}
\textbf{Input:} They want to create a post on the college's equal opportunities committee to ensure that their opinions can be aired freely. \\
\vspace{0.5em}
\textbf{Output:} Sie wollen einen Posten im Ausschuss f\"ur Chancengleichheit des Kollegiums einrichten, um sicherzustellen, dass ihre Meinungen frei zur Sprache gebracht werden k\"onnen. \\
\vspace{0.5em}
\textbf{Parallel decode (uniform loss):}
\end{flushleft}
\vspace{-1.5em}
\begin{center}
\resizebox{\textwidth}{!}{%
\begin{tikzpicture}[every node/.style={anchor=base,inner sep=0.1em}]
\node[gray!60] (node-0-0) {Sie\_};
\node[gray!60,base right=0cm of node-0-0] (node-0-1) {wollen\_};
\node[gray!60,base right=0cm of node-0-1] (node-0-2) {einen\_};
\node[gray!60,base right=0cm of node-0-2] (node-0-3) {Posten\_};
\node[gray!60,base right=0cm of node-0-3] (node-0-4) {im\_};
\node[gray!60,base right=0cm of node-0-4] (node-0-5) {Ausschuss\_};
\node[gray!60,base right=0cm of node-0-5] (node-0-6) {f\"ur\_};
\node[gray!60,base right=0cm of node-0-6] (node-0-7) {Chancengleichheit\_};
\node[gray!60,base right=0cm of node-0-7] (node-0-8) {des\_};
\node[gray!60,base right=0cm of node-0-8] (node-0-9) {Koll};
\node[gray!60,base right=0cm of node-0-9] (node-0-10) {egi};
\node[gray!60,base right=0cm of node-0-10] (node-0-11) {ums\_};
\node[gray!60,base right=0cm of node-0-11] (node-0-12) {einrichten\_};
\node[blue,base right=0cm of node-0-12] (node-0-13) {\uline{, \_}};
\node[gray!60,base right=0cm of node-0-13] (node-0-14) {um\_};
\node[gray!60,base right=0cm of node-0-14] (node-0-15) {sicherzustellen\_};
\node[gray!60,base right=0cm of node-0-15] (node-0-16) {, \_};
\node[gray!60,base right=0cm of node-0-16] (node-0-17) {dass\_};
\node[gray!60,base right=0cm of node-0-17] (node-0-18) {ihre\_};
\node[gray!60,base right=0cm of node-0-18] (node-0-19) {Mein};
\node[gray!60,base right=0cm of node-0-19] (node-0-20) {ungen\_};
\node[gray!60,base right=0cm of node-0-20] (node-0-21) {frei\_};
\node[gray!60,base right=0cm of node-0-21] (node-0-22) {zur\_};
\node[gray!60,base right=0cm of node-0-22] (node-0-23) {Sprache\_};
\node[gray!60,base right=0cm of node-0-23] (node-0-24) {gebracht\_};
\node[gray!60,base right=0cm of node-0-24] (node-0-25) {werden\_};
\node[gray!60,base right=0cm of node-0-25] (node-0-26) {k\"onnen\_};
\node[gray!60,base right=0cm of node-0-26] (node-0-27) {.\_};
\node[gray!60,below=0.1cm of node-0-0] (node-1-0) {Sie\_};
\node[gray!60,base right=0cm of node-1-0] (node-1-1) {wollen\_};
\node[blue,base right=0cm of node-1-1] (node-1-2) {\uline{einen\_}};
\node[gray!60,base right=0cm of node-1-2] (node-1-3) {Posten\_};
\node[gray!60,base right=0cm of node-1-3] (node-1-4) {im\_};
\node[gray!60,base right=0cm of node-1-4] (node-1-5) {Ausschuss\_};
\node[gray!60,base right=0cm of node-1-5] (node-1-6) {f\"ur\_};
\node[gray!60,base right=0cm of node-1-6] (node-1-7) {Chancengleichheit\_};
\node[gray!60,base right=0cm of node-1-7] (node-1-8) {des\_};
\node[gray!60,base right=0cm of node-1-8] (node-1-9) {Koll};
\node[gray!60,base right=0cm of node-1-9] (node-1-10) {egi};
\node[gray!60,base right=0cm of node-1-10] (node-1-11) {ums\_};
\node[gray!60,base right=0cm of node-1-11] (node-1-12) {einrichten\_};
\node[black,base right=0cm of node-1-12] (node-1-13) {, \_};
\node[gray!60,base right=0cm of node-1-13] (node-1-14) {um\_};
\node[gray!60,base right=0cm of node-1-14] (node-1-15) {sicherzustellen\_};
\node[gray!60,base right=0cm of node-1-15] (node-1-16) {, \_};
\node[gray!60,base right=0cm of node-1-16] (node-1-17) {dass\_};
\node[gray!60,base right=0cm of node-1-17] (node-1-18) {ihre\_};
\node[blue,base right=0cm of node-1-18] (node-1-19) {\uline{Mein}};
\node[gray!60,base right=0cm of node-1-19] (node-1-20) {ungen\_};
\node[gray!60,base right=0cm of node-1-20] (node-1-21) {frei\_};
\node[gray!60,base right=0cm of node-1-21] (node-1-22) {zur\_};
\node[gray!60,base right=0cm of node-1-22] (node-1-23) {Sprache\_};
\node[gray!60,base right=0cm of node-1-23] (node-1-24) {gebracht\_};
\node[gray!60,base right=0cm of node-1-24] (node-1-25) {werden\_};
\node[gray!60,base right=0cm of node-1-25] (node-1-26) {k\"onnen\_};
\node[gray!60,base right=0cm of node-1-26] (node-1-27) {.\_};
\node[gray!60,below=0.1cm of node-1-0] (node-2-0) {Sie\_};
\node[blue,base right=0cm of node-2-0] (node-2-1) {\uline{wollen\_}};
\node[black,base right=0cm of node-2-1] (node-2-2) {einen\_};
\node[gray!60,base right=0cm of node-2-2] (node-2-3) {Posten\_};
\node[gray!60,base right=0cm of node-2-3] (node-2-4) {im\_};
\node[gray!60,base right=0cm of node-2-4] (node-2-5) {Ausschuss\_};
\node[gray!60,base right=0cm of node-2-5] (node-2-6) {f\"ur\_};
\node[gray!60,base right=0cm of node-2-6] (node-2-7) {Chancengleichheit\_};
\node[blue,base right=0cm of node-2-7] (node-2-8) {\uline{des\_}};
\node[gray!60,base right=0cm of node-2-8] (node-2-9) {Koll};
\node[gray!60,base right=0cm of node-2-9] (node-2-10) {egi};
\node[gray!60,base right=0cm of node-2-10] (node-2-11) {ums\_};
\node[gray!60,base right=0cm of node-2-11] (node-2-12) {einrichten\_};
\node[black,base right=0cm of node-2-12] (node-2-13) {, \_};
\node[gray!60,base right=0cm of node-2-13] (node-2-14) {um\_};
\node[gray!60,base right=0cm of node-2-14] (node-2-15) {sicherzustellen\_};
\node[gray!60,base right=0cm of node-2-15] (node-2-16) {, \_};
\node[blue,base right=0cm of node-2-16] (node-2-17) {\uline{dass\_}};
\node[gray!60,base right=0cm of node-2-17] (node-2-18) {ihre\_};
\node[black,base right=0cm of node-2-18] (node-2-19) {Mein};
\node[gray!60,base right=0cm of node-2-19] (node-2-20) {ungen\_};
\node[gray!60,base right=0cm of node-2-20] (node-2-21) {frei\_};
\node[gray!60,base right=0cm of node-2-21] (node-2-22) {zur\_};
\node[gray!60,base right=0cm of node-2-22] (node-2-23) {Sprache\_};
\node[gray!60,base right=0cm of node-2-23] (node-2-24) {gebracht\_};
\node[gray!60,base right=0cm of node-2-24] (node-2-25) {werden\_};
\node[gray!60,base right=0cm of node-2-25] (node-2-26) {k\"onnen\_};
\node[blue,base right=0cm of node-2-26] (node-2-27) {\uline{.\_}};
\node[blue,below=0.1cm of node-2-0] (node-3-0) {\uline{Sie\_}};
\node[black,base right=0cm of node-3-0] (node-3-1) {wollen\_};
\node[black,base right=0cm of node-3-1] (node-3-2) {einen\_};
\node[gray!60,base right=0cm of node-3-2] (node-3-3) {Posten\_};
\node[gray!60,base right=0cm of node-3-3] (node-3-4) {im\_};
\node[gray!60,base right=0cm of node-3-4] (node-3-5) {Ausschuss\_};
\node[gray!60,base right=0cm of node-3-5] (node-3-6) {f\"ur\_};
\node[blue,base right=0cm of node-3-6] (node-3-7) {\uline{Chancengleichheit\_}};
\node[black,base right=0cm of node-3-7] (node-3-8) {des\_};
\node[gray!60,base right=0cm of node-3-8] (node-3-9) {Koll};
\node[gray!60,base right=0cm of node-3-9] (node-3-10) {egi};
\node[blue,base right=0cm of node-3-10] (node-3-11) {\uline{ums\_}};
\node[gray!60,base right=0cm of node-3-11] (node-3-12) {einrichten\_};
\node[black,base right=0cm of node-3-12] (node-3-13) {, \_};
\node[gray!60,base right=0cm of node-3-13] (node-3-14) {um\_};
\node[blue,base right=0cm of node-3-14] (node-3-15) {\uline{sicherzustellen\_}};
\node[gray!60,base right=0cm of node-3-15] (node-3-16) {, \_};
\node[black,base right=0cm of node-3-16] (node-3-17) {dass\_};
\node[blue,base right=0cm of node-3-17] (node-3-18) {\uline{ihre\_}};
\node[black,base right=0cm of node-3-18] (node-3-19) {Mein};
\node[blue,base right=0cm of node-3-19] (node-3-20) {\uline{ungen\_}};
\node[gray!60,base right=0cm of node-3-20] (node-3-21) {frei\_};
\node[gray!60,base right=0cm of node-3-21] (node-3-22) {zur\_};
\node[gray!60,base right=0cm of node-3-22] (node-3-23) {Sprache\_};
\node[gray!60,base right=0cm of node-3-23] (node-3-24) {gebracht\_};
\node[gray!60,base right=0cm of node-3-24] (node-3-25) {werden\_};
\node[gray!60,base right=0cm of node-3-25] (node-3-26) {k\"onnen\_};
\node[black,base right=0cm of node-3-26] (node-3-27) {.\_};
\node[black,below=0.1cm of node-3-0] (node-4-0) {Sie\_};
\node[black,base right=0cm of node-4-0] (node-4-1) {wollen\_};
\node[black,base right=0cm of node-4-1] (node-4-2) {einen\_};
\node[gray!60,base right=0cm of node-4-2] (node-4-3) {Posten\_};
\node[gray!60,base right=0cm of node-4-3] (node-4-4) {im\_};
\node[gray!60,base right=0cm of node-4-4] (node-4-5) {Ausschuss\_};
\node[blue,base right=0cm of node-4-5] (node-4-6) {\uline{f\"ur\_}};
\node[black,base right=0cm of node-4-6] (node-4-7) {Chancengleichheit\_};
\node[black,base right=0cm of node-4-7] (node-4-8) {des\_};
\node[gray!60,base right=0cm of node-4-8] (node-4-9) {Koll};
\node[blue,base right=0cm of node-4-9] (node-4-10) {\uline{egi}};
\node[black,base right=0cm of node-4-10] (node-4-11) {ums\_};
\node[blue,base right=0cm of node-4-11] (node-4-12) {\uline{einrichten\_}};
\node[black,base right=0cm of node-4-12] (node-4-13) {, \_};
\node[blue,base right=0cm of node-4-13] (node-4-14) {\uline{um\_}};
\node[black,base right=0cm of node-4-14] (node-4-15) {sicherzustellen\_};
\node[blue,base right=0cm of node-4-15] (node-4-16) {\uline{, \_}};
\node[black,base right=0cm of node-4-16] (node-4-17) {dass\_};
\node[black,base right=0cm of node-4-17] (node-4-18) {ihre\_};
\node[black,base right=0cm of node-4-18] (node-4-19) {Mein};
\node[black,base right=0cm of node-4-19] (node-4-20) {ungen\_};
\node[gray!60,base right=0cm of node-4-20] (node-4-21) {frei\_};
\node[gray!60,base right=0cm of node-4-21] (node-4-22) {zur\_};
\node[gray!60,base right=0cm of node-4-22] (node-4-23) {Sprache\_};
\node[gray!60,base right=0cm of node-4-23] (node-4-24) {gebracht\_};
\node[blue,base right=0cm of node-4-24] (node-4-25) {\uline{werden\_}};
\node[gray!60,base right=0cm of node-4-25] (node-4-26) {k\"onnen\_};
\node[black,base right=0cm of node-4-26] (node-4-27) {.\_};
\node[black,below=0.1cm of node-4-0] (node-5-0) {Sie\_};
\node[black,base right=0cm of node-5-0] (node-5-1) {wollen\_};
\node[black,base right=0cm of node-5-1] (node-5-2) {einen\_};
\node[gray!60,base right=0cm of node-5-2] (node-5-3) {Posten\_};
\node[gray!60,base right=0cm of node-5-3] (node-5-4) {im\_};
\node[blue,base right=0cm of node-5-4] (node-5-5) {\uline{Ausschuss\_}};
\node[black,base right=0cm of node-5-5] (node-5-6) {f\"ur\_};
\node[black,base right=0cm of node-5-6] (node-5-7) {Chancengleichheit\_};
\node[black,base right=0cm of node-5-7] (node-5-8) {des\_};
\node[blue,base right=0cm of node-5-8] (node-5-9) {\uline{Koll}};
\node[black,base right=0cm of node-5-9] (node-5-10) {egi};
\node[black,base right=0cm of node-5-10] (node-5-11) {ums\_};
\node[black,base right=0cm of node-5-11] (node-5-12) {einrichten\_};
\node[black,base right=0cm of node-5-12] (node-5-13) {, \_};
\node[black,base right=0cm of node-5-13] (node-5-14) {um\_};
\node[black,base right=0cm of node-5-14] (node-5-15) {sicherzustellen\_};
\node[black,base right=0cm of node-5-15] (node-5-16) {, \_};
\node[black,base right=0cm of node-5-16] (node-5-17) {dass\_};
\node[black,base right=0cm of node-5-17] (node-5-18) {ihre\_};
\node[black,base right=0cm of node-5-18] (node-5-19) {Mein};
\node[black,base right=0cm of node-5-19] (node-5-20) {ungen\_};
\node[blue,base right=0cm of node-5-20] (node-5-21) {\uline{frei\_}};
\node[gray!60,base right=0cm of node-5-21] (node-5-22) {zur\_};
\node[gray!60,base right=0cm of node-5-22] (node-5-23) {Sprache\_};
\node[gray!60,base right=0cm of node-5-23] (node-5-24) {gebracht\_};
\node[black,base right=0cm of node-5-24] (node-5-25) {werden\_};
\node[blue,base right=0cm of node-5-25] (node-5-26) {\uline{k\"onnen\_}};
\node[black,base right=0cm of node-5-26] (node-5-27) {.\_};
\node[black,below=0.1cm of node-5-0] (node-6-0) {Sie\_};
\node[black,base right=0cm of node-6-0] (node-6-1) {wollen\_};
\node[black,base right=0cm of node-6-1] (node-6-2) {einen\_};
\node[blue,base right=0cm of node-6-2] (node-6-3) {\uline{Posten\_}};
\node[gray!60,base right=0cm of node-6-3] (node-6-4) {im\_};
\node[black,base right=0cm of node-6-4] (node-6-5) {Ausschuss\_};
\node[black,base right=0cm of node-6-5] (node-6-6) {f\"ur\_};
\node[black,base right=0cm of node-6-6] (node-6-7) {Chancengleichheit\_};
\node[black,base right=0cm of node-6-7] (node-6-8) {des\_};
\node[black,base right=0cm of node-6-8] (node-6-9) {Koll};
\node[black,base right=0cm of node-6-9] (node-6-10) {egi};
\node[black,base right=0cm of node-6-10] (node-6-11) {ums\_};
\node[black,base right=0cm of node-6-11] (node-6-12) {einrichten\_};
\node[black,base right=0cm of node-6-12] (node-6-13) {, \_};
\node[black,base right=0cm of node-6-13] (node-6-14) {um\_};
\node[black,base right=0cm of node-6-14] (node-6-15) {sicherzustellen\_};
\node[black,base right=0cm of node-6-15] (node-6-16) {, \_};
\node[black,base right=0cm of node-6-16] (node-6-17) {dass\_};
\node[black,base right=0cm of node-6-17] (node-6-18) {ihre\_};
\node[black,base right=0cm of node-6-18] (node-6-19) {Mein};
\node[black,base right=0cm of node-6-19] (node-6-20) {ungen\_};
\node[black,base right=0cm of node-6-20] (node-6-21) {frei\_};
\node[gray!60,base right=0cm of node-6-21] (node-6-22) {zur\_};
\node[gray!60,base right=0cm of node-6-22] (node-6-23) {Sprache\_};
\node[blue,base right=0cm of node-6-23] (node-6-24) {\uline{gebracht\_}};
\node[black,base right=0cm of node-6-24] (node-6-25) {werden\_};
\node[black,base right=0cm of node-6-25] (node-6-26) {k\"onnen\_};
\node[black,base right=0cm of node-6-26] (node-6-27) {.\_};
\node[black,below=0.1cm of node-6-0] (node-7-0) {Sie\_};
\node[black,base right=0cm of node-7-0] (node-7-1) {wollen\_};
\node[black,base right=0cm of node-7-1] (node-7-2) {einen\_};
\node[black,base right=0cm of node-7-2] (node-7-3) {Posten\_};
\node[blue,base right=0cm of node-7-3] (node-7-4) {\uline{im\_}};
\node[black,base right=0cm of node-7-4] (node-7-5) {Ausschuss\_};
\node[black,base right=0cm of node-7-5] (node-7-6) {f\"ur\_};
\node[black,base right=0cm of node-7-6] (node-7-7) {Chancengleichheit\_};
\node[black,base right=0cm of node-7-7] (node-7-8) {des\_};
\node[black,base right=0cm of node-7-8] (node-7-9) {Koll};
\node[black,base right=0cm of node-7-9] (node-7-10) {egi};
\node[black,base right=0cm of node-7-10] (node-7-11) {ums\_};
\node[black,base right=0cm of node-7-11] (node-7-12) {einrichten\_};
\node[black,base right=0cm of node-7-12] (node-7-13) {, \_};
\node[black,base right=0cm of node-7-13] (node-7-14) {um\_};
\node[black,base right=0cm of node-7-14] (node-7-15) {sicherzustellen\_};
\node[black,base right=0cm of node-7-15] (node-7-16) {, \_};
\node[black,base right=0cm of node-7-16] (node-7-17) {dass\_};
\node[black,base right=0cm of node-7-17] (node-7-18) {ihre\_};
\node[black,base right=0cm of node-7-18] (node-7-19) {Mein};
\node[black,base right=0cm of node-7-19] (node-7-20) {ungen\_};
\node[black,base right=0cm of node-7-20] (node-7-21) {frei\_};
\node[blue,base right=0cm of node-7-21] (node-7-22) {\uline{zur\_}};
\node[gray!60,base right=0cm of node-7-22] (node-7-23) {Sprache\_};
\node[black,base right=0cm of node-7-23] (node-7-24) {gebracht\_};
\node[black,base right=0cm of node-7-24] (node-7-25) {werden\_};
\node[black,base right=0cm of node-7-25] (node-7-26) {k\"onnen\_};
\node[black,base right=0cm of node-7-26] (node-7-27) {.\_};
\node[black,below=0.1cm of node-7-0] (node-8-0) {Sie\_};
\node[black,base right=0cm of node-8-0] (node-8-1) {wollen\_};
\node[black,base right=0cm of node-8-1] (node-8-2) {einen\_};
\node[black,base right=0cm of node-8-2] (node-8-3) {Posten\_};
\node[black,base right=0cm of node-8-3] (node-8-4) {im\_};
\node[black,base right=0cm of node-8-4] (node-8-5) {Ausschuss\_};
\node[black,base right=0cm of node-8-5] (node-8-6) {f\"ur\_};
\node[black,base right=0cm of node-8-6] (node-8-7) {Chancengleichheit\_};
\node[black,base right=0cm of node-8-7] (node-8-8) {des\_};
\node[black,base right=0cm of node-8-8] (node-8-9) {Koll};
\node[black,base right=0cm of node-8-9] (node-8-10) {egi};
\node[black,base right=0cm of node-8-10] (node-8-11) {ums\_};
\node[black,base right=0cm of node-8-11] (node-8-12) {einrichten\_};
\node[black,base right=0cm of node-8-12] (node-8-13) {, \_};
\node[black,base right=0cm of node-8-13] (node-8-14) {um\_};
\node[black,base right=0cm of node-8-14] (node-8-15) {sicherzustellen\_};
\node[black,base right=0cm of node-8-15] (node-8-16) {, \_};
\node[black,base right=0cm of node-8-16] (node-8-17) {dass\_};
\node[black,base right=0cm of node-8-17] (node-8-18) {ihre\_};
\node[black,base right=0cm of node-8-18] (node-8-19) {Mein};
\node[black,base right=0cm of node-8-19] (node-8-20) {ungen\_};
\node[black,base right=0cm of node-8-20] (node-8-21) {frei\_};
\node[black,base right=0cm of node-8-21] (node-8-22) {zur\_};
\node[blue,base right=0cm of node-8-22] (node-8-23) {\uline{Sprache\_}};
\node[black,base right=0cm of node-8-23] (node-8-24) {gebracht\_};
\node[black,base right=0cm of node-8-24] (node-8-25) {werden\_};
\node[black,base right=0cm of node-8-25] (node-8-26) {k\"onnen\_};
\node[black,base right=0cm of node-8-26] (node-8-27) {.\_};
\end{tikzpicture}
}
\end{center}
\vspace{-1em}
\caption{Example decodes using models trained with the binary tree loss and the uniform loss. Within each row, the underlined blue words are those being inserted, and the gray words represent those from the final output that have not yet been generated. We observe that both models are able to achieve a high degree of parallelism, with the binary tree model matching the logarithmic theoretical lower bound on the number of parallel decoding iterations thanks to its training objective.}
\label{fig:example-parallel}
\end{figure*}

We also perform a more careful analysis of the extent of the parallelism achieved by our highest-scoring models. In Figure~\ref{fig:plot-parallel}, we plot the number of decoding iterations taken vs.\ the output length $n$ for each development sentence. We also plot the theoretical lower bound of $\lfloor \log_2 n \rfloor + 1$ and the upper bound of $n$ on the number of iterations. Note that greedy decoding takes $n$ steps by definition. Our best model comes impressively close to the lower bound across the entire development set, rarely deviating by more than 1 or 2 iterations. This demonstrates that our framework is capable of producing high-quality output using a sub-linear (i.e.\ logarithmic) number of generation steps.

\subsection{Test Results}

Finally we report results in Table~\ref{tab:results-newstest2014} on the newstest2014 test set using our best hyperparameters as measured on the development set. When compared with related approaches, we find that we match the high quality of models requiring a linear number of iterations while using a logarithmic number of generation steps. In practice, as shown in Figure~\ref{fig:plot-parallel}, we rarely require more than 10 generation steps, meaning our empirical complexity even matches that of \citet{lee-emnlp-2018} who use a constant 10 steps. When trained with the binary tree loss, we find that the Insertion Transformer is able to match the standard Transformer model while requiring substantially fewer generation iterations.

\begin{table}[t!]
\centering
\resizebox{\columnwidth}{!}{%
\begin{tabular}{lcc}
\toprule
\bfseries Model & \bfseries BLEU & \bfseries Iterations \\
\midrule
Autoregressive Left-to-Right \\
\quad Transformer \cite{vaswani-nips-2017} & 27.3 & $n$ \\
\midrule
Semi-Autoregressive Left-to-Right \\
\quad SAT \cite{wang-emnlp-2018} & 24.83 & $n / 6$ \\
\quad Blockwise Parallel \cite{stern-neurips-2018} & 27.40 & $\approx n / 5$ \\
\midrule
Non-Autoregressive \\
\quad NAT \cite{gu-iclr-2018} & 17.69 & $1$ \\
\quad Iterative Refinement \cite{lee-emnlp-2018} & 21.61 & $10$ \\
\midrule
Our Approach (Greedy) \\
\quad Insertion Transformer + Left-to-Right & 23.94 & $n$ \\
\quad Insertion Transformer + Binary Tree  & 27.29 & $n$ \\
\quad Insertion Transformer + Uniform  & 27.12 & $n$ \\
Our Approach (Parallel) \\
\quad Insertion Transformer + Binary Tree & 27.41 & $\approx \log_2 n$ \\
\quad Insertion Transformer + Uniform  & 26.72 & $\approx \log_2 n$ \\
\bottomrule
\end{tabular}
}
\caption{BLEU scores on the newstest2014 test set for the WMT 2014 English-German translation task. Our parallel decoding strategy attains the same level of accuracy reached by linear-complexity models while using only a logarithmic number of decoding iterations.}
\label{tab:results-newstest2014}
%\vspace{-1.5em}
\end{table}
\section{Related Work}
There has been prior work on non-left-to-right autoregressive generation. \citet{vinyals-iclr-2015} explores the modeling of sets, where generation order does not matter. \citet{ford-emnlp-2018} explores language modeling where select words (i.e., functional words) are generated first, and the rest are filled in using a two-pass process.
There has also been prior work in hierarchical autoregressive image generation \cite{reed-icml-2017}, where $\log n$ steps are required to generate $n$ tokens. This bears some similarity to our balanced binary tree order.

\citet{shah-aaai-2018} also recently proposed generating language with a dynamic canvas. Their work can be seen as a continuous relaxation version of our model, wherein their canvas is an embedding space, while our canvas contains discrete tokens. They applied their approach to language modeling tasks, whereas we apply ours to conditional language generation in machine translation.

In addition, there has been recent work on non-autoregressive machine translation \cite{gu-iclr-2018,lee-emnlp-2018} and semi-autoregressive translation \cite{stern-neurips-2018,wang-emnlp-2018}.
The key difference between our work and prior work is that the Insertion Transformer framework can accommodate for a dynamically growing canvas size while still achieving sub-linear generation complexity. Other models also tend to degrade with increasing parallelism, while our model trained with the balanced binary tree loss suffers no model degradation under parallel decoding.

We must also mention the concurrent work of \citet{gu-arxiv-2019}.\footnote{We note that the research described in this paper was carried out independently and without communication with \citet{gu-arxiv-2019} and \citet{welleck-arxiv-2019}.\label{independent}} They similarly use an insertion-based framework to generate sequences, but there are some differences and tradeoffs between our approaches.
The main difference is that we model each successive canvas explicitly after a set of insertions, while \citet{gu-arxiv-2019} model the canvas implicitly by conditioning on the insertion sequence. Consequently, \citet{gu-arxiv-2019}'s approach is autoregressive, can rely on cached decoder states, and permits standard beam search, while our approach must recompute the decoder states with each iteration, but is partially autoregressive and thereby allows for parallel decoding.
% The main difference being that the autoregressive approach permits beam-search decoding, while our partially autoregressive approach permits parallel decoding; consequently, \citet{gu-arxiv-2019} can rely on cached decoder states while we must recompute the decoder states with each iteration.
\citet{gu-arxiv-2019} also explored tree-based orders, but while they found the syntactic tree order from a dependency parser to do slightly worse than a left-to-right baseline, we find our balanced binary tree approach to match the standard Transformer even when using parallel decoding.

Finally, we also note that \citet{welleck-arxiv-2019}\footnotemark[1] concurrently explored generation using a tree formulation, similar to our Insertion Transformer implementation. However, they did not explore the balanced binary tree policy examined in this work, nor did they adapt their model for parallel generation, instead opting to use a serialized in-order traversal. Moreover, on a machine translation task, \citet{welleck-arxiv-2019} found left-to-right generation to be superior to their learned orderings, while our balanced binary tree approach is able to match the performance of the standard Transformer.

\section{Conclusion}
In this paper, we presented the Insertion Transformer, a partially autoregressive model for sequence generation based on insertion operations. Our model can be trained to follow arbitrary generation orderings, such as a left-to-right order or a balanced binary tree order, or can be optimized to learn all possible orderings, making it also applicable to completion or infilling tasks. The model can be decoded serially, producing one token at a time, or it can be decoded in parallel with simultaneous insertions at multiple locations. When using the binary tree loss, we find empirically that we can generate sequences of length $n$ using close to the asymptomatic limit of $\lfloor \log_2 n \rfloor + 1$ steps without any quality degradation. This allows us to match the performance of the standard Transformer on the WMT 2014 English-German translation task while using substantially fewer iterations during decoding.

\section*{Acknowledgements}
We give thanks to Zhifeng Chen, Geoffrey Hinton, Mohammad Norouzi, Oscar T{\"a}ckstr{\"o}m and the Google Brain Team for useful comments, discussions, and technical assistance.

\bibliography{paper}
\bibliographystyle{icml2019}

%\appendix
%\section{Do \emph{not} have an appendix here}
%
%\textbf{\emph{Do not put content after the references.}}
%
%Put anything that you might normally include after the references in a separate
%supplementary file.

%We recommend that you build supplementary material in a separate document.
%If you must create one PDF and cut it up, please be careful to use a tool that
%doesn't alter the margins, and that doesn't aggressively rewrite the PDF file.
%pdftk usually works fine. 

%\textbf{Please do not use Apple's preview to cut off supplementary material.} In
%previous years it has altered margins, and created headaches at the camera-ready
%stage. 
%%%%%%%%%%%%%%%%%%%%%%%%%%%%%%%%%%%%%%%%%%%%%%%%%%%%%%%%%%%%%%%%%%%%%%%%%%%%%%%
%%%%%%%%%%%%%%%%%%%%%%%%%%%%%%%%%%%%%%%%%%%%%%%%%%%%%%%%%%%%%%%%%%%%%%%%%%%%%%%

\end{document}